# 基于条件生成对抗网络的咬翼片语义分割

蒋芸；谭宁；张海；彭婷婷

**摘要**：目前，对于咬翼片 X 射线图像进行语义分割的研究非常具有挑战性，研究的重点在于将其分割成龋齿、牙釉质、牙本质、牙髓、牙冠、修复体和牙根管等类型。现阶段对咬翼片进行语义分割的主要方法是 U 型深度卷积神经网络，但其存在准确率偏低的问题。为了提高对咬翼片语义分割的准确率，本文将条件生成对抗网络（cGAN）和 U 型网络结构（U-Net）相结合对咬翼片进行语义分割。实验结果表明，cGAN 结合 U-Net 的准确率达到 69.7%，比之前使用 U 型卷积神经网络的准确率 56.4%提高了 13.3%。

**关键字**：生成对抗网络；语义分割；深度学习；U 型网络；对抗学习

## Bitewing Radiography Semantic Segmentation Base on Conditional Generative Adversarial Nets

JiangYun；TanNing；ZhangHai；PengTingting

【**Abstract**】 Currently, Segmentation of bitewing radiograpy images is a very challenging task. The focus of the study is to segment it into caries, enamel, dentin, pulp, crowns, restoration and root canal treatments. The main method of semantic segmentation of bitewing radiograpy images at this stage is the U-shaped deep convolution neural network, but its accuracy is low. in order to improve the accuracy of semantic segmentation of bitewing radiograpy images, this paper proposes the use of Conditional Generative Adversarial network (cGAN) combined with U-shaped network structure (U-Net) approach to semantic segmentation of bitewing radiograpy images. The experimental results show that the accuracy of cGAN combined with U-Net is 69.7%, which is 13.3% higher than the accuracy of u-shaped deep convolution neural network of 56.4%.

【**Key words**】Generative Adversarial Nets（GAN）；Semantic Segmentation；deep leaning； U-Nets；Adversarial Leaning

## 1 前言

  语义分割[1]是医学图像分析的一个重要组成部分，是图像理解的基石性技术，在自动驾驶技术中对于街道场景的辨别和理解、无人机关于着陆点的判断以及穿戴设备中都有着

非常重要的应用。图像由许多的像素组成，而语义分割是根据图片中不同的内容表达不同的语义对这些像素用不同的颜色进行标注，这与深度学习[2-4]领域的突破是高度相关的。近些年来，无监督学习已经成为了研究的热点，变分自编码器[5]、生成对抗网络（Generative Adversarial Nets，GAN）[18]等无监督模型受到越来越多的关注。在人工智能高速发展的时代，GAN 的提出不仅满足了相关领域的研究和应用需求，也带来了新的发展动力。特别是在图像和视觉领域中对 GAN 的研究和应用是最为广泛，已经可以通过随机数字生成人脸、从低分辨率图像生成高分辨率图像等[6]、电脑病毒监控[7]、自然语言和语音处理[8-9]、棋类竞赛程序[10]等。此外，GAN 已经开始逐渐应用到医学图像处理中：从单一的术前磁共振图像直接生成病人特定的超声探头诱导前列腺运动模型[11]、模拟超声探头的空间位置上有条件地对解剖学精确的图像进行采样[12]、检测恶性前列腺癌[13]等问题的研究中。

本文的工作主要是通过识别咬翼片 X 射线图像中的龋齿对其进行语义分割，由于龋齿会破坏牙齿结构传播细菌性疾病，牙医主要根据咬翼片 X 射线图像来诊断和治疗龋齿。自动化龋齿病变检测技术为牙科医生提供潜在的诊断数据，并有助于识别各种疾病的迹象[14]。目前，牙科鉴定过程是手工进行的，这使得其非常耗时且主观性差。为了解决这个问题，研究一种自动化分割咬翼片 X 线图像的方法[15-17]，在 2004 年 Jain 和 H.Chen[15]提出了一种使用像素的积分投影来检测牙齿之间的间隙的分割方法。2005 年，Zhou 和 Abdel-Mottaleb[16] 提出了一种牙齿自动分割方法通过使用主动轮廓法的方法提取牙齿的轮廓。P.L. Lin 等[17]在 2013 年，建立了一种自动且高效的牙齿隔离方法，能够实现上下颌分离，单齿隔离，过分割验证和下分割检测。牙齿隔离对于计算机辅助牙科诊断和牙齿自动识别的一个非常重要的步骤，因为牙齿隔离将直接影响特征提取和分类的准确性。Olaf Ronneberger 等在 2015 年，使用 U-Net 对咬翼片进行语义分割，将咬翼片 X 射线图像分割成龋齿、牙釉质、牙本质、牙髓、牙冠、修复体和牙根管等 7 个部分，但是分割的准确率偏低，为了解决这个问题，于是我们在上述研究的基础上引入条件生成对抗神经网络对咬翼片 X 射线进行语义分割，以提高分割的准确率和精度。

## 2 相关理论

自从 2014 年 Ian Goodfellow 提出生成对抗网络[18]的概念后，生成对抗网络受到了越来越多研究者们广泛的关注，近几年，生成对抗神经网络成为无监督[19]复杂概率分布学

习最流行的方法，因为生成器(G)和判别器(D)可以采用当前最热门的深度神经网络[20]。Yann LeCun 曾评价 GAN 是"进十年来机器学习领域最酷的技术"，并在 2016 巴塞罗那的 NIPS 上提到将"预测学习（Predictive Learning）"落脚到 GAN，本节主要介绍文中用的模型所依赖的 *GAN*[18]、*cGAN*[24]原理。

**2.1 生成对抗神经网络**

生成对抗神经网络(*GAN*)[18]由两个神经网络组成：包含一个生成器(*G*)和一个判别器(*D*)。生成器(*G*)的目的是生成近似真实数据分布 $p_{data}$ 的样本来欺骗判别器(*D*)，使判别器(*D*)无法区分输入的数据来自真实数据 $p_{data}$ 还是生成器(*G*)；判别器(*D*)的目的是正确的区分数据样本来自真实的数据分布 $p_{data}$ 还是生成器(*G*)，当输入的数据来自真实的数据分布 $p_{data}$ 时，判别器(*D*)的目标是使输出的概率 $D(x) \approx 1$，当输入的数据来自生成器(*G*)时，判别器(*D*)的目标是使 $D(G(z)) \approx 0$，同时生成器(*G*)的目标是使得 $D(G(z)) \approx 1$。两个网络在一个极小-极大的游戏里相互迭代，相互竞争优化各自的网络参数，使目标函数达到那什均衡[21]，收敛时，我们期望 $p_{data} = p_g$，其中 $p_g$ 是生成器(*G*)生成的数据样本分布。本文使用下面的目标函数来优化生成器的参数($\theta_G$)和判别器的参数($\theta_D$)：

$$\max_D \min_G (\theta_G, \theta_D) = \mathbb{E}_{x \sim p_{data}}[\log D(x)] + \mathbb{E}_{z \sim p_z(z)}[\log(1 - D(G(z)))] \quad (1)$$

其中 *x* 取自于真实的数据分布 $p_{data}$，*z* 取自于先验分布 $p_z(z)$ (例如正态分布)，$\mathbb{E}(\cdot)$ 表示计算期望值。

**2.2 条件生成对抗网络**

条件生成对抗网络(*cGAN*)[24]是生成器(*G*)和判别器(*D*)都输入一个额外的信息 *y* 作为条件，*y* 可以是任何类型的辅助信息（例如手写体，*y* 可以表示为数字的类别）。额外辅助信息 *y* 与样本数据 *x* 进行拼接后输入到生成器(*G*)和判别器(*D*)，使生成对抗网络扩展为了条件模型，目标函数变为如下：

$$\min_G \max_D \mathcal{L}(D, G) = \mathbb{E}_{x \sim p_{data}(x)}[\log D(x|y)] + \mathbb{E}_{z \sim p_z(z)}[\log(1 - D(G(x|y)))] \quad (2)$$

## 3 生成对抗网络对咬翼片的语义分割

### 3.1 损失函数

生成对抗神经网络(GAN)[18]的生成模型需要学习到一个从随机噪声 z 到输出图像 y 之间的映射关系[22]，$G: z \rightarrow y$。条件生成对抗网络(cGAN)[24]通过学习到图片 x 和随机噪声 z 与图片 y 之间的映射，$G:\{x, z\} \rightarrow y$。

实验中所用的条件生成对抗神经网络的目标函数描述如下：

$$\mathcal{L}_{cGAN}(G, D) = \mathbb{E}_{x \sim pdata(x), y}[\log D(x, y)] + \mathbb{E}_{x \sim pdata(x), z \sim pz(z)}[\log(1 - D(x, G(x, z)))] \quad (3)$$

x 为输入的图片样本数据，y 是通过人工标注的标准分割图像，生成器(G)尝试最小化目标函数，判别器(D)尝试最大化目标函数：$G^* = \min_G \max_D \mathcal{L}_{cGAN}(G, D)$。

通过实验发现将 cGAN 的目标函数与传统的损失函数(像 $L_1$ 距离)相结合分割后的效果会更好，$L_1$ 的距离函数如下：

$$\mathcal{L}_{L1}(G) = E_{x \sim pdata(x), y, z \sim pz(z)}[|||y - G(x, z)|||] \quad (4)$$

其中 y 是语义分割的目标图像，$G(x, z)$ 是训练样本数据输入生成器中生成的语义分割图像。最终目标函数变为如下所示：

$$G^* = \arg\min_G \max_D \mathcal{L}_{cGAN}(G, D) + \lambda \mathcal{L}_{L1}(G) \quad (5)$$

由于 $L_1$ 距离函数会产生模糊化效果，通过加入超参 $\lambda$ 进行控制。当生成器能够对输入的咬翼片 X 射线图像进行准确的分割时，对参数的改变非常的敏感，适当的选择超参 $\lambda$ 的值是非常的重要，推荐超参的值 $\lambda = 100$ 比较合理。

### 3.2 模型结构

生成器(G)的训练过程分两部分：（1）输入一张咬翼片 X 射线图像到生成器(G)，生成器(G)输出一张语义分割后的图像，比较生成器(G)输出的分割图像和人工标注的标准分割图像之间的误差，通过误差调整生成器(G)的权重。（2）输入一对假图像（咬翼片 X 射线图像，生成器通过该咬翼片生成的分割图像）到判别器(D)中，由于生成器(G)是生成一张无限接近于目标的分割图像，使判别器(D)误认为生成器(G)输出的分割图像是人工标注的标注图像，期待判别器输出的结果为"1"。比较判别器(D)输出的结果与标准的正确结果"1"之间的误差，从而优化生成器(G)的权重，使其生成的分割图像更加接近目标分割图像。

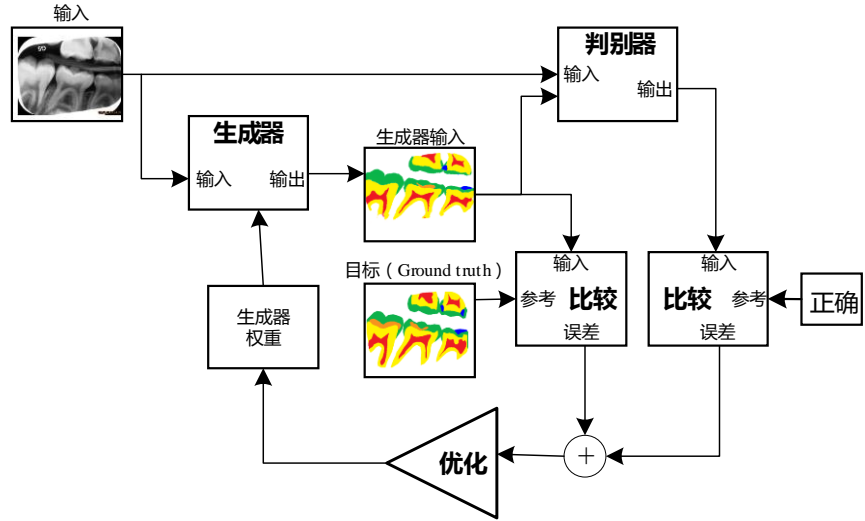

图 1　生成器(*G*)训练的详细过程

判别器(*D*)的训练过程同样分两部分：（1）输入一对真实的图像对（咬翼片 X 射线图像，人工标注的标准目标分割图像）到判别器(*D*)，判别器(*D*)输出判别后的结果，由于已知输入的是真实的数据样本，所以判别器(*D*)输出的结果应该为"1"，通过比较判别器(*D*)输出的结果与标准答案"1"之间的差值，然后优化判别器(*D*)的权重。（2）同理，右边输入一对假图像对（咬翼片 X 射线图像，生成器输出的分割图像）到判别器(*D*)中，已知输入的数据样本为假，所以判别器(*D*)输出的结果应该为"0"，比较判别器(*D*)输出的结果与标准答案"0"之间的差值，然后调整判别器(*D*)的权重，使判别器(*D*)能够正确的区分图像是人工标注的标准的分割图像还是由生成器(*G*)生成的分割图像。

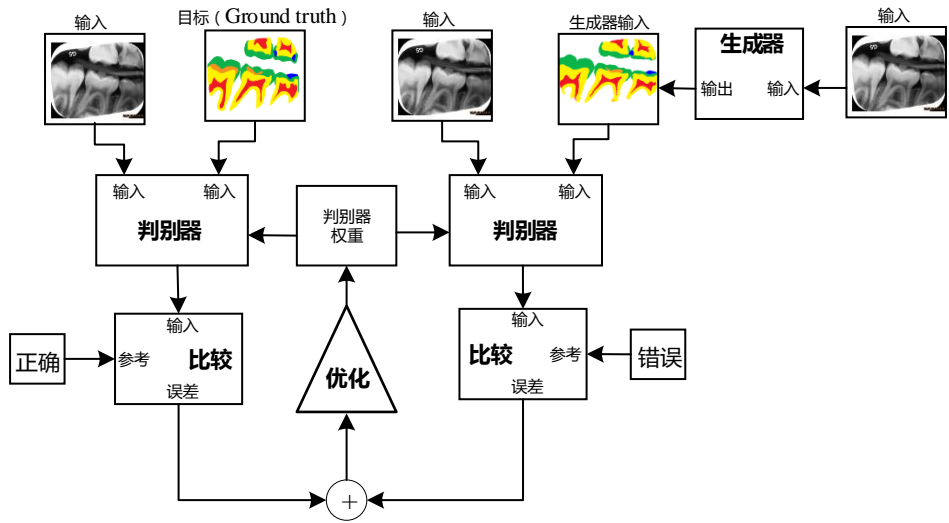

图 2　判别器(*D*)训练的详细过程

## 3.3 数据增强

由于只能提供少量咬翼片 X 射线样本图像对网络结构进行训练，但深度神经网络大量参数的训练需要大量的训练样本，这造成了数据的饥饿性，我们通过数据增强的方式对训练样本进行扩充从而解决这个问题。数据的增强对于提高网络分类的准确率、防止过拟合、和网络的鲁棒性至关重要。主要通过对图像的旋转、水平翻转、垂直翻转、平移变换、图像灰度值的变化等方法对训练数据进行增强。

## 3.4 网络结构

本论文工作基于深度卷积对抗神经网络(*DCGAN*)[25]和条件生成对抗神经网络(*cGAN*)[24]进行实现，使用 *DCGAN* 中推荐的训练参数进行训练，训练时使用 *Adam* 优化算法[26]（$\beta_1 = 0.5$，$\beta_2=0.999$，$\varepsilon=10^{-8}$），学习率为 $lr = 0.0002$，$mini-batch=2$，每层输出的结果进行批量归一化从而减少每层之间的依赖性，提高各网络层之间的独立性[27-28]，训练 20 个周期，输出的分割图像的大小为 $256×256$，生成器(*G*)和判别器(*D*)具体的网络结构如下所示：

表 1：生成器的网络结构

| 操作 | 卷积核大小 | 步伐 | 卷积核数 | 归一化 | 激活函数 | 备注 |
|---|---|---|---|---|---|---|
| e 1：卷积 | 5×5 | 2×2 | 64 | 是 | Leaky ReLU | - |
| e 2：卷积 | 5×5 | 2×2 | 128 | 是 | Leaky ReLU | - |
| e 3：卷积 | 5×5 | 2×2 | 256 | 是 | Leaky ReLU | - |
| e 4：卷积 | 5×5 | 2×2 | 512 | 是 | Leaky ReLU | - |
| e 5：卷积 | 5×5 | 2×2 | 512 | 是 | Leaky ReLU | - |
| e 6：卷积 | 5×5 | 2×2 | 512 | 是 | Leaky ReLU | - |
| e 7：卷积 | 5×5 | 2×2 | 512 | 是 | Leaky ReLU | - |
| e 8：卷积 | 5×5 | 2×2 | 512 | 是 | Leaky ReLU | - |
| d 1：反卷积 | 5×5 | 2×2 | 512+512 | 是 | ReLU | 连接[d1,e7] dropout:0.5 |
| d 2：反卷积 | 5×5 | 2×2 | 512+512 | 是 | ReLU | 连接[d2,e6] dropout:0.5 |
| d 3：反卷积 | 5×5 | 2×2 | 512+512 | 是 | ReLU | 连接[d3,e5] dropout:0.5 |
| d 4：反卷积 | 5×5 | 2×2 | 512+512 | 是 | ReLU | 连接[d4,e4] |
| d 5：反卷积 | 5×5 | 2×2 | 256+256 | 是 | ReLU | 连接[d5,e3] |
| d 6：反卷积 | 5×5 | 2×2 | 128+128 | 是 | ReLU | 连接[d6,e2] |
| d 7：反卷积 | 5×5 | 2×2 | 64+64 | 是 | ReLU | 连接[d7,e1] |
| d 8：反卷积 | 5×5 | 2×2 | 3 | - | ReLU | - |
| 全连接 | - | - | 3 | - | tanh | - |

表 2：判别器的网络结构

| 操作 | 卷积核大小 | 步伐 | 卷积核数 | 批量归一化 | 激活函数 |
|---|---|---|---|---|---|
| h 0：卷积 | 5×5 | 2×2 | 64 | 是 | Leaky ReLU |
| h 1：卷积 | 5×5 | 2×2 | 128 | 是 | Leaky ReLU |
| h 2：卷积 | 5×5 | 2×2 | 256 | 是 | Leaky ReLU |
| h 3：卷积 | 5×5 | 2×2 | 512 | 是 | Leaky ReLU |
| h 4：全连接 | - | - | 16*16*512 | 否 | - |
| 全连接 | - | - | 1 | 否 | Sigmoid |

生成器中使用 $U-net$ 网络结构[23]，在 $d_1, d_2, d_3$ 三层使用 $dropout$，每层随机删除 50% 的节点以防止过拟合。

以前许多分割方法的网络结构使用编码器-解码器($Encoder-Decoder$)[29]，这种网络结构是通过向下采样，逐渐降低采样层，直到达到一个瓶颈层，将提取的信息变为一个一维向量；在这一点上过程被逆转，逐渐向上采样，最后还原成图像。这个网络结构要求所有的信息流通过所有的网络层，包括瓶颈层。在许多图像翻译问题中，输入和输出之间共享大量低级信息，但希望这些信息直接穿过网络。

为了使生成器($G$)能够避免出现这种信息的瓶颈，本文使用了跳远连接(skip connection)，遵循 $U-net$ 网络结构[23]，具体操作是将网络的第 $i$ 层和网络的第 $n-i$ 进行连接跳远连接，每个跳远连接只是简单的将第 $i$ 层网络输出的所有通道和第 $n-i$ 层的所有输出进行连接（n 为网络结构的总层数），作为第 $n-i+1$ 层节点的输入。

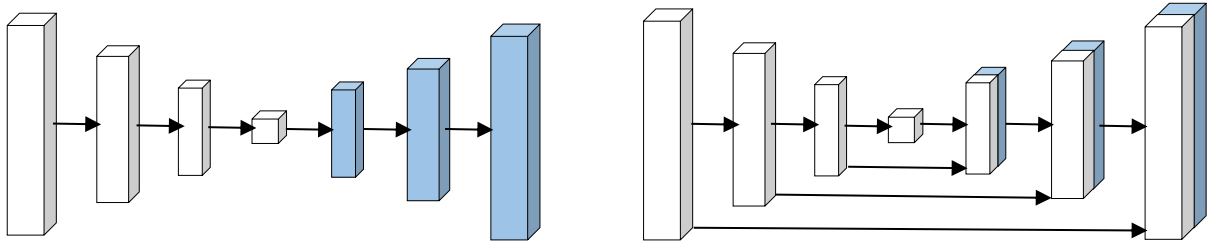

图 3 生成器的两种网络结构，编码器-解码器($Encoder-Decoder$)结构（左边）；使用跳远连接的 $U-net$ 网络结构（右边）

## 4 实验结果分析

实验的目标是为计算机自动检测龋齿研究出一种合适的自动进行语义分割的方法，用于识别七种类型的区域，包括龋齿、牙釉质、牙本质、牙髓、牙冠、牙根管和修复区域，并用不同的颜色进行标记。图 4（a）显示了一张咬翼片 X 射线的图像，图 4（b）为该图像由医生进行手工标记的 7 个类型区域。

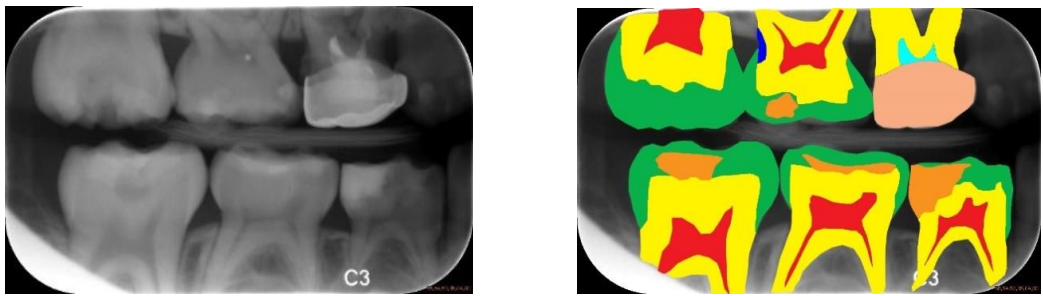

（a）原始牙科图像　　　　　　　（b）7种颜色标记的咬翼片图像

图 4 （a）咬翼片图像和（b）颜色代表的区域：龋齿（蓝色）、牙釉质（绿色）、牙本质（黄色）、牙髓（红色）、牙冠（肤色）、修复（橙色）、牙根管（青色）

训练的数据样本来自使用 80 个患者的咬翼片 X 射线图像和医生手工标记的标准分割图像对，由于图像的分辨率不统一，因此把图像重新进行缩放到256×256像素，并且归一化灰度值到[–1,1]的范围。通过数据增强的方法把训练数据集扩充到 28800 对，使用这些数据集训练我们的模型，最后训练出来的模型对咬翼片 X 射线图像的分割效果如图 5 所示，可以看到对于常见类型牙釉质、牙本质，牙髓能够很好的进行分割。

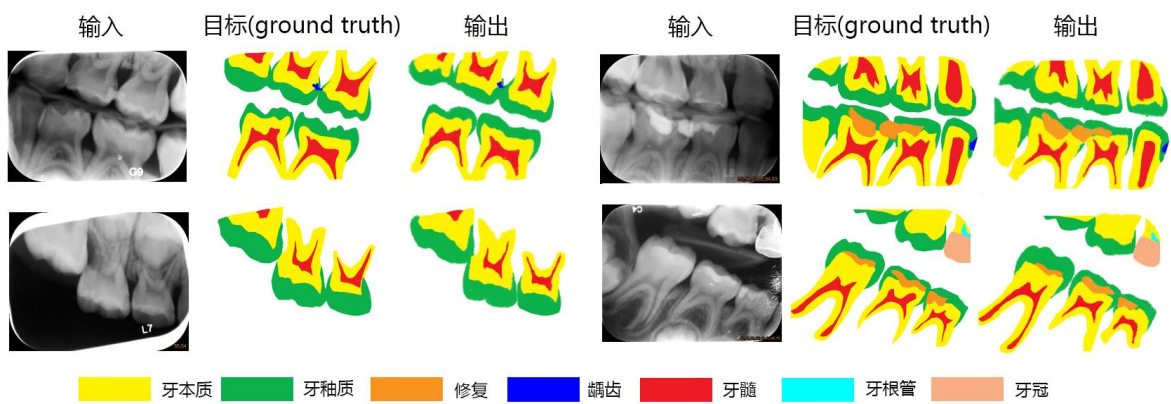

图 5　使用 cGAN+U-Net 对咬翼片 X 射线语义分割的结果

## 4.1 结果评估

主要通过评估模型的精确度（Precision）、准确性（Accuracy）：计算 True Positives（$TP$）和 True Negatives（$TN$）、相似性（Dice Similarity）这三个主要标准来评估所提出的方法的性能。

训练出来的模型对测试集（40 个患者的咬翼片 X 射线图像）进行语义分割后，通过模型分割出来的 7 个部位与医生手工标记的标准结果进行对比（表 3），可以看出对最常见的类别（牙釉质、牙本质、牙髓）分割的相似($D$)度超过 75%，对龋齿、牙冠、牙根管等其他部位的分割效果还需要进一步的提高。

表 3　测试集上不同部位分割的平均结果

| 部位 | P | TP | TN | D |
| --- | --- | --- | --- | --- |
| 龋齿 | 0.418 | 0.768 | 0.973 | 0.584 |
| 牙釉质 | 0.588 | 0.886 | 0.923 | 0.759 |
| 牙本质 | 0.641 | 0.797 | 0.848 | 0.781 |
| 牙髓 | 0.542 | 0.878 | 0.941 | 0.751 |
| 牙冠 | 0.494 | 0.790 | 0.901 | 0.567 |
| 修复 | 0.513 | 0.792 | 0.946 | 0.663 |
| 牙根管 | 0.433 | 0.721 | 0.958 | 0.597 |

模型通过 U-Net 和 cGAN 结合后，语义分割的相似度($D$)达到 69.7%，而文献[30]采用 U-Net 分割的相似度($D$)最高为 56.4%，相对来说提高了 13.3%，表 4 为两个模型在几个评估标准上的比较。

表 4　测试集两个模型的平均结果

| 模型 | P | TP | TN | D |
| --- | --- | --- | --- | --- |
| U-Net[30] | 0.453 | 0.613 | 0.983 | 0.564 |
| cGAN +U-Net | **0.546** | **0.784** | 0.956 | **0.697** |

实验结果表明，条件生成对抗网络对于图像的语义分割是很有前景的。模型对牙齿常见的类别（牙本质、牙釉质、牙髓）分割比较准确，由于数据增强的原因，有大量的训练样本可以对这些部位进行训练；而对于龋齿，牙根管等这些部位分割的准确率还比较低，据我们所知道，假如有更多的训练数据来学习这些类，分割的准确率将会更高。标准的目标分割图像中有时牙齿并未标记出来，加大了模型训练的难度，假设提供的训练数据标注更加严格，模型的整体性能可能会更好。

## 5　总结

本文通过使用条件生成对抗网络结合 U 型网络结构对咬翼片 X 射线图像进行语义分割，将其分割成龋齿、牙冠、牙釉质、牙本质、牙髓、牙根管和修复体等七种类型的区域。最终实验结果表明，我们所用的模型对咬翼片语义分割的准确率与 U 型深度卷积神经网络分割的准确率相比，有了明显提高，特别是对于常见类别（牙釉质、牙本质、牙髓）分割的准确率超过了 75%。但是，如果包含龋齿、牙根管等类型的训练数据样本增多，模型识别的准确率和性能是否会进一步的提高，还需要进一步实验测试。

# 参考文献

[1] J. Long, E. Shelhamer, and T. Darrell. Fully convolutional networks for semantic segmentation. In CVPR, pages 3431–3440, 2015.

[2] HINTON G, OSINDERO S, THE Y.A fast learning algorithm for deep belief nets[J]. Neural Computation, 2006, 18(7):1527-1554

[3] Hinton G E, Salakhutdinov R R. Reducing the dimensionality of data with neural networks. Science, 2006, 313(5786):504−507

[4] Le Cun Y, Bengio Y, Hinton G. Deep learning. Nature, 2015,521(7553): 436−444

[5] Kingma, Diederik P and Welling, Max. Auto-encoding variational bayes. International Conference on Learning Representations (2014), 2013.

[6] Ledig C, Theis L, Husz ár F, Caballero J, Cunningham A, Acosta A, Aitken A, Tejani A, Totz J, Wang Z H, Shi W Z. Photo-realistic single image super-resolution using a generative adversarial network. ar Xiv preprint ar Xiv: 1609.04802, 2016.

[7] Hu WW, Tan Y. Generating adversarial malware examples for black-box attacks based on GAN. ar Xiv preprint ar Xiv: 1702.05983, 2017.

[8] Li J W, Monroe W, Shi T L, Jean S, Ritter A, Jurafsky D. Adversarial learning for neural dialogue generation. ar Xiv preprint ar Xiv: 1701.06547, 2017.

[9] Yu L T, Zhang W N, Wang J, Yu Y. Seq GAN: sequence generative adversarial nets with policy gradient. ar Xiv preprint ar Xiv: 1609.05473, 2016.

[10] Chidambaram M, Qi Y J. Style transfer generative adversarial networks: learning to play chess differently. ar Xiv preprint ar Xiv: 1702.06762, 2017.

[11] Yipeng Hu1,2, Eli Gibson1, Tom Vercauteren1, Hashim U. Ahmed3. Intraoperative Organ Motion Models with an Ensemble of Conditional Generative Adversarial Networks.in MICCAI, 2017.

[12] Yipeng Hu1,2, Eli Gibson1, Li-Lin Lee3. Freehand Ultrasound Image Simulation with Spatially Conditioned Generative Adversarial Networks. in MICCAI, 2017

[13] Simon Kohl, David Bonekamp, Heinz-Peter Schlemmer, Kaneschka Yaqubi. Adversarial Networks for the Detection of Aggressive Prostate Cancer. in MICCAI, 2017

[14] P.L. Tan, R.W. Evans, and M.V. Morgan, "Caries, bitewings, and treatment decisions," Australian Dental Journal, vol. 47, no. 2, pp. 138-141, 2002.

[15] A. Jain and H. Chen, "Matching of dental X-ray images for human identification," Pattern Recognition, vol. 37, pp. 1519-1532, 2004.

[16] J. Zhou and M. Abdel-Mottaleb, "A content-based system for human identification based on bitewing dental X-ray images," Pattern Recognition, vol. 38, no. 11, pp. 2132-2142, 2005.

[17] P.L. Lin, P.W. Huang, Y.S. Cho, and C.H. Kuo, "An automatic and effective tooth isolation method for dental radiographs," Opto-Electronics Review, vol. 21, no. 1, pp. 126-136, 2013.

[18] Goodfellow, I. J., Pouget-Abadie, J., Mirza, M., Xu, B., Warde-Farley, D., Ozair, S., Courville, A., and Bengio, Y. (2014). Generative adversarial nets. In NIPS'2014.

[19] Ilya Sutskever, Rafal Jozefowicz, Karol Gregor, et al. Towards principled unsupervised learning. arXiv preprint arXiv:1511.06440, 2015.

[20] Goodfellow I, Bengio Y, Courville A. Deep Learning. Cambridge, UK: MIT Press, 2016.

[21] Ratliff L J, Burden S A, Sastry S S. Characterization and computation of local Nash equilibria in continuous games. In: Proceedings of the 51st Annual Allerton Conference on Communication, Control, and Computing (Allerton). Monticello, IL, USA: IEEE, 2013. 917−924

[22] Zhiwu Huang, Bernhard Kratzwald, Danda Pani Paudel, Jiqing Wu, Luc Van Gool. Face Translation between Images and Videos using Identity-aware CycleGAN. arXiv preprint arXiv:1712.00971,2017.12

[23] O. Ronneberger, P. Fischer, and T. Brox. U-net: Convolutional networks for biomedical image segmentation. In MICCAI, pages 234–241. Springer, 2015. 2, 3

[24] M. Mirza and S. Osindero. Conditional generative adversarial nets. arXiv preprint arXiv:1411.1784, 2014. 2

[25] A. Radford, L. Metz, and S. Chintala. Unsupervised representation learning with deep convolutional generative adversarial networks. arXiv preprint arXiv:1511.06434, 2015.

[26] D. Kingma and J. Ba. Adam: A method for stochastic optimization. ICLR, 2015.

[27] T. Salimans, I. Goodfellow, W. Zaremba, V. Cheung, A. Radford, and X. Chen. Improved techniques for training gans. In Advances in Neural Information Processing Systems, pages 2226–2234, 2016.

[28] Ishaan Gulrajani, Faruk Ahmed, Martin Arjovsky, Vincent Dumoulin, Aaron Courville. Improved Training of Wasserstein GANs. NIPS, 2017

[29] D. Pathak, P. Krahenbuhl, J. Donahue, T. Darrell, and A. A. Efros. Context encoders: Feature learning by inpainting. CVPR, 2016.

[30] Olaf Ronneberger, Philipp Fischer and Thomas Brox. Dental X-ray image segmentation using a U-shaped Deep Convolutional network. ISBI 2015